\documentclass{article} 
\usepackage{iclr2027_conference,times}

\usepackage{amsmath,amsfonts,bm}

\def\eqref#1{equation~\ref{#1}}

\def\1{\bm{1}}

\DeclareMathAlphabet{\mathsfit}{\encodingdefault}{\sfdefault}{m}{sl}
\SetMathAlphabet{\mathsfit}{bold}{\encodingdefault}{\sfdefault}{bx}{n}

\usepackage{hyperref}
\usepackage{url}
\usepackage{graphicx}
\usepackage{booktabs}
\usepackage[table]{xcolor}
\usepackage{multirow}
\usepackage{algorithm}
\usepackage{algpseudocode}
\usepackage{xcolor}
\usepackage{bm}
\usepackage{pifont}

\newcommand{\cmark}{\ding{51}}  
\newcommand{\xmark}{\ding{55}}  

\definecolor{myred}{RGB}{220,0,0}       
\definecolor{mygreen}{RGB}{0,210,0}     
\definecolor{myblue}{RGB}{0,0,210}     
\definecolor{myorange}{RGB}{200,80,0}   

\algrenewcommand\algorithmicrequire{\textbf{Input:}}
\algrenewcommand\algorithmicensure{\textbf{Output:}}
\title{Learning Adaptive and Visually Aligned Conditions for Generative Zero-Shot Learning}

\author{
Haojie Pu\thanks{Equal contribution.} \, \textsuperscript{1},
Zhuoming Li\footnotemark[1] \, \textsuperscript{1},
Yongbiao Gao\textsuperscript{2},
Hui Liu\textsuperscript{3},
Junhui Hou\textsuperscript{4},
Yuheng Jia\thanks{Corresponding author.} \, \textsuperscript{1}\\
\normalfont
\textsuperscript{1}Southeast University
\quad
\textsuperscript{2}Qilu University of Technology
\\
\normalfont
\textsuperscript{3}Saint Francis University
\quad
\textsuperscript{4}City University of Hong Kong
}
\iclrArxivCopy
\begin{document}

\maketitle

\begin{abstract}
Generative zero-shot learning (ZSL) synthesizes visual features for unseen classes by learning a semantic-conditioned generator from seen classes.
Since semantic conditions determine the knowledge transfer from seen to unseen classes, their effectiveness is critical for generating plausible features. 
However, existing class-level semantic prototypes lack semantic diversity to capture intra-class variations, while exhibiting inconsistent inter-class structures with the visual space due to the semantic–visual gap. 
To address these issues, we propose Adaptive Attribute Distribution and Visual Structure Alignment (AAVS), a unified semantic condition learning framework for generative ZSL. 
Specifically, the Adaptive Attribute Distribution (AAD) learns dimension-adaptive attribute distributions with discriminability-guided
variation calibration to capture transferable semantic diversity, while Visual Structure Alignment (VSA) aligns the diverse attributes with refined visual prototypes to learn visual inter-class structures. 
By jointly modeling semantic diversity and visual structural consistency, AAVS provides more effective generation conditions for unseen classes to generate visual features. 
Extensive quantitative and qualitative evaluations on three benchmarks demonstrate that AAVS outperforms existing generative ZSL methods and validate the effectiveness of the learned generation conditions.
\end{abstract}

\section{Introduction}
\label{Introduction}

Conventional visual recognition relies on a closed-set assumption: every test class must be represented in the training data, which is difficult to satisfy in real-world applications, where new classes may emerge without labeled data.
Zero-shot learning (ZSL) \citep{ZSL09,ZSL19} offers a promising solution by leveraging semantic information to transfer knowledge learned from seen classes to unseen classes, enabling recognition without visual examples.
With the development of generative models, including variational autoencoders (VAEs) \citep{VAE13}, generative adversarial networks (GANs) \citep{GAN14}, and diffusion models \citep{DMs,DDGAN22}, generative zero-shot learning (Generative ZSL) \citep{f-CLSWGAN18} further learns a semantic-conditioned generator from seen classes and uses it to synthesize visual features for unseen classes, which has consequently become an effective paradigm to ZSL.

Since the generator can only be trained on seen classes, the semantic conditions play a crucial role in transferring its generation capability from seen to unseen classes to synthesize plausible features for unseen classes. Existing methods typically adopt class-level semantic conditions, such as predefined attributes \citep{VSBoost23,DSP23,VADS24,ViFR25,RLVC26} or text embeddings \citep{16CVPR,SHIP23,GenZSL25}, which build an effective bridge for knowledge transfer. however, two inherent yet usually overlooked Challenges remain:
\begin{figure*}[t]
    \centering
    \includegraphics[width=\textwidth]{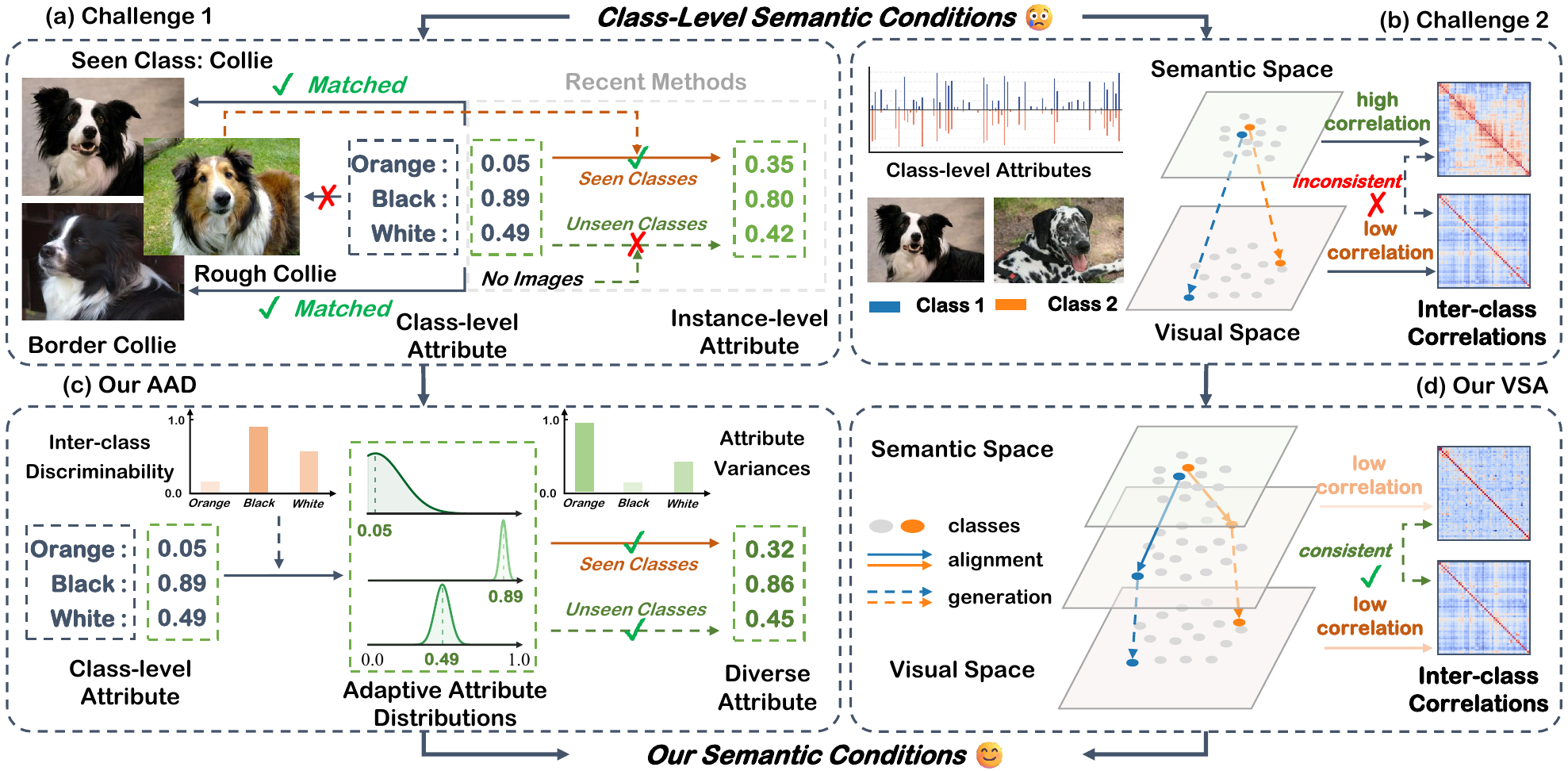}
    \setlength{\abovecaptionskip}{0pt}
    \setlength{\belowcaptionskip}{-2pt}
    \caption{ 
Challenges of class-level semantic conditions and our motivation.
(a) Challenge 1 \textcolor{myred}{\xmark}: Class-level semantics lack diversity to describe intra-class variations. 
Although recent methods introduce visual-guided semantic enhancement, they rely on seen visual features and are hard to generalize to unseen classes . 
(b) Challenge 2 \textcolor{myred}{\xmark}: Semantically similar classes may exhibit distinct visual appearances (left), revealing the semantic-visual gap and inconsistent inter-class structures between two spaces (right). 
(c) Our AAD \textcolor{mygreen}{\cmark}: Different attributes contribute differently to class discrimination, motivating attribute-adaptive distribution learning to obtain diverse semantic conditions without visual features. 
(d) Our VSA \textcolor{mygreen}{\cmark}: Semantic representations are aligned with visual structures to obtain visually consistent generation conditions.
    }
    \label{fig:motivation}
\end{figure*}

\textbf{1) Static semantic conditions fail to accurately describe intra-class instance variations.}
As shown in Fig.~\ref{fig:motivation}(a), the class-level attributes of Collie assign a high value to \textbf{Black} and a low value to \textbf{Orange}, which well describes the black-and-white Border Collie but fails to represent the warm-orange Rough Collie.
Consequently, instances within the same class share identical semantic conditions, leaving the generator without instance-varying semantic guidance to capture intra-class variations.
This limits the learning of accurate visual–semantic correspondence \citep{ZeroDiff25} and results in insufficient diversity of generated features.
Recent methods attempt to alleviate this issue by leveraging visual features from seen classes to obtain instance-level semantic enhancements during generator training \citep{DSP23,VADS24,IAB24}. However, since visual features of unseen classes are unavailable, such visual-guided semantic enhancements are hard to obtain diverse semantics for unseen classes and still rely on a single semantic condition.

\textbf{2) Class-level semantic conditions exhibit inconsistent inter-class structures with the real visual space.}
As shown in Fig.~\ref{fig:motivation}(b), two classes with nearly identical attributes but exhibit markedly different visual appearances. Such mismatch reflects the semantic–visual gap \cite{DSP23,ViFR25}, which further leads to discrepancies in inter-class structures between two domains, as illustrated by the correlation matrices in Fig.~\ref{fig:motivation}(c). Since generative ZSL relies on semantic conditions to capture inter-class structures and learn cross-domain mappings, this structural inconsistency makes knowledge transfer from seen to unseen classes more challenging, causing generated features to deviate from the real visual distribution.

These observations raise a fundamental question: \textbf{what properties should an effective semantic condition possess for generative ZSL?} We answer this question from two perspectives:

\textbf{1) Semantic diversity.} 
Class-level semantic conditions fail to describe intra-class instance variations due to their limited semantic diversity. 
Since visual features within the same class exhibit rich variations as shown in Fig.~\ref{fig:motivation}(a), semantic conditions should also provide a variation space to characterize potential instance-level semantics. 
Furthermore, different attributes contribute differently to class discrimination. 
As shown in Fig.~\ref{fig:motivation}(c), \textbf{Black} and \textbf{White} is more discriminative for Collie than \textbf{Orange}, indicating that different attributes play different roles in representing class semantics and variations. 
Motivated by this observation, we propose \textbf{Adaptive Attribute Distribution (AAD)}, which learns adaptive attribute distributions for different attribute dimensions and adjusts their variation ranges according to attribute discriminability. 
By learning entirely from the shared semantic space, AAD can naturally transfer to unseen classes and provide diverse semantic conditions for feature synthesis.

\textbf{2) Structural consistency.} 
Due to the semantic-visual gap, the inter-class structures encoded by class-level semantic conditions are inconsistent with the visual space, as shown in Fig.~\ref{fig:motivation}(b). 
Since generators tend to inherit the structures from conditioning spaces, such discrepancies can be propagated to generated features, causing them to deviate from the real visual distribution. 
To obtain semantic conditions with visual structural consistency, we propose \textbf{Visual Structure Alignment (VSA)}, which projects attributes into the visual space and aligns the embeddings with the refined visual prototypes to learn visual inter-class structures. 
As shown in Fig.~\ref{fig:motivation}(d), the aligned attribute embeddings exhibit more consistent inter-class structures with the visual space, thereby enabling more reliable cross-domain feature generation and mitigating the distribution shift between generated and real features.

To this end, we propose \textbf{AAVS}, a unified semantic condition learning framework consisting of Adaptive Attribute Distribution (AAD) and Visual Structure Alignment (VSA). Starting from class-level semantic conditions, AAVS learns semantic representations with both semantic diversity and visual structural consistency. These representations jointly serve as generation conditions, enabling the generator to establish more accurate visual-semantic correspondence. Since AAVS is learned solely from the shared attribute space, it can be effectively transferred to unseen classes and generate plausible visual features that better match the real visual distribution.

Our contributions are summarized as follows:
\begin{itemize}
   \item To address the limitation that class-level semantic conditions cannot describe instance variations, we propose \textbf{Adaptive Attribute Distribution}, which constructs adaptive attribute distributions to obtain semantically diverse attributes.
   \item To address the class-level semantic conditions' inconsistent inter-class structures with the real visual space, we propose \textbf{Visual Structure Alignment}, which projects attributes into the visual space to obtain attribute embeddings with visual structural consistency.
   \item Extensive experiments on widely used benchmarks demonstrate that our method outperforms state-of-the-arts and can obtain effective generation conditions.
\end{itemize}

\section{Related Work}
\label{Related Work}
\subsection{Generative Zero-Shot Learning}
Generative Zero-shot Learning synthesizes visual features for unseen classes by using conditional generative models, such as GANs \citep{GAN14}, VAEs \citep{VAE13} and Diffusion Models \citep{DMs}. 
Focusing on improving generated feature quality, existing methods are mainly divided into two lines: 
(1) Visual feature enhancement. 
Some approaches \citep{TF-VAEGAN, FREE21,ViFR25,ZeroDiff25} improve visual feature by fine-tuning the backbone or adapting pretrained visual features, while other approaches \citep{ICCE22, VSBoost23} further exploit visual-semantic alignment to obtain semantically guided visual representations.
(2) Semantic condition enhancement.
Early methods explore richer semantic representations, such as text embeddings \cite{16CVPR} and CLIP-based representations \cite{SHIP23,GenZSL25}, whereas recent approaches attempt to evolve semantic prototypes under the guidance of visual information \citep{DSP23,VADS24,ZeroDiff25}.
Despite the progress, existing semantic enhancement relies on visual information, making it challenging to transfer to unseen classes where visual samples are unavailable. In contrast, our Adaptive Attribute Distribution directly learns transferable attribute variation patterns and obtains diverse semantic conditions for unseen classes.

\subsection{Embedding-based Zero-Shot Learning}
Embedding-based zero-shot learning represents another mainstream paradigm for ZSL, where the semantic-visual gap remains a central challenge.
Early methods \citep{LearningDeepEmbedding17,LantentEmbeddings16} learn a shared embedding space to align visual features with semantic representations, and recent methods \citep{TransZero++23,ZeroViT24,SVIP25} further leverage attention mechanisms to capture discriminative regions. 
In addition, semantic-guided methods \citep{VSPCN25,AENet25, PSVMA+25,SAGE26} exploit attribute prototypes to enhance semantic-aware representation learning.
In generative settings, semantic information directly controls visual feature synthesis, and the semantic-visual gap may cause the generated features to inherit incorrect inter-class structures.
To address this issue, our Visual Structure Alignment incorporates visual structures into generation conditions before synthesis, enabling conditions that better reflect visual relationships.

\subsection{Effective Generation Condition}
Generative models learn to synthesize samples that follow the real data distribution \citep{GAN14}.
VAEs and GAN-based models introduce latent noise sampled from Gaussian priors to capture stochastic variations \citep{VAE13,GAN14,LatentGAN18,VAEGAN18}. 
However, latent noise alone provide limited guidance for modeling complex high-dimensional distributions, resulting in incomplete mode coverage and limited diversity \citep{UnrolledGAN17,WassersteinGAN17,LatentGAN18}. 
To enhance controllability, high-level semantic information is incorporated as generation conditions to provide semantic constraints and capture inter-class relationships \citep{ConditionalGAN14,f-CLSWGAN18}. 
Nevertheless, semantic-visual structural inconsistency may cause generated samples to deviate from the real visual distribution \citep{DSP23,VADS24,ViFR25}. 
These observations emphasize the need for generation conditions that captures intra-class variations while preserving visual-consistent inter-class structures, motivating us to propose AAVS, which learns adaptive attribute distributions and aligns semantic embeddings with visual structures to construct more effective generation conditions.

\section{Methodology}
\label{Method}

\begin{figure*}[t]
    \centering
    \includegraphics[width=\textwidth]{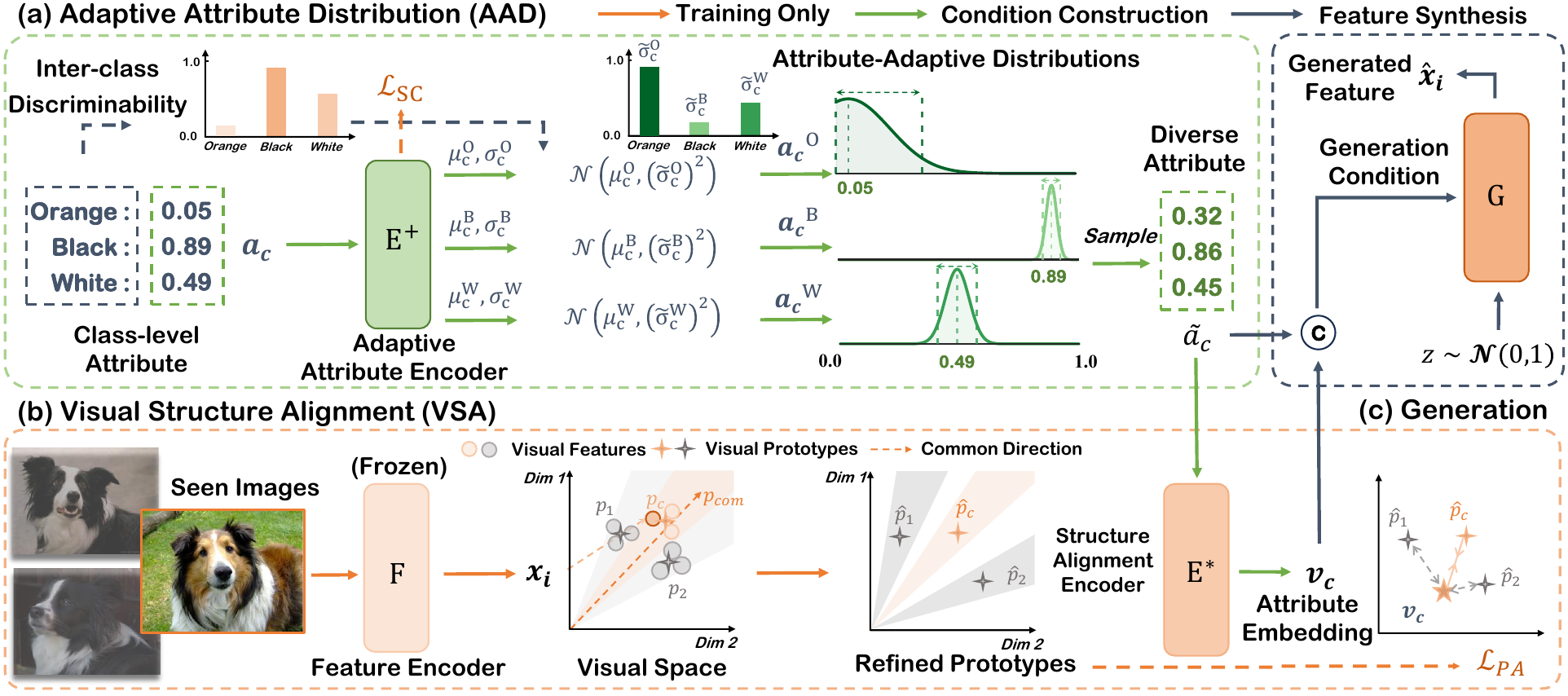}
    \setlength{\abovecaptionskip}{0pt}
    \setlength{\belowcaptionskip}{-2pt}
    \caption{ 
      Overview of AAVS.
       Given class attributes, AAVS constructs generation conditions via  \textcolor{mygreen}{$\bm{\rightarrow}$}: (a) Adaptive Attribute Distribution (AAD), learns adaptive attribute distributions with inter-class discriminability calibrating the variation magnitude and semantic consistency constraint $\mathcal{L}_{SC}$, producing diverse attributes $\tilde{a}_c$. 
       And (b) Visual Structure Alignment (VSA),
       aggregates seen features into prototypes and refines them by removing dominant common directions (\textcolor{myorange}{$\bm{\rightarrow}$}), which supervise the attribute embedding $v_c$ through prototype alignment loss $\mathcal{L}_{PA}$ to incorporate  visual class structures.
       In the (c) generation phase ($\bm{\rightarrow}$), the resulting $a_c^{+}$ and $v_c$ are concatenated as generation conditions, with latent noise $z$ to synthesize visual features.
    }
    \label{fig:model_pipeline}
\end{figure*}

\subsection{Notations}
Generative ZSL learns a semantic-conditioned generator from seen classes to synthesize unseen-class visual features. The seen and unseen datasets are defined as $\mathcal{D}^{s}=\{(x_i^s,y_i^s)\}_{i=1}^{N^s}$ and $\mathcal{D}^{u}=\{(x_i^u,y_i^u)\}_{i=1}^{N^u}$, where $\mathcal{Y}^{s}\cap\mathcal{Y}^{u}=\emptyset$ and unseen visual features are unavailable during training. Following \citet{ZSL19}, seen datsets are split into training and test sets. Each class $c\in\mathcal{Y}^{s}\cup\mathcal{Y}^{u}$ is associated with a semantic attribute $a_c\in\mathbb{R}^{|A|}$. Conventional ZSL (CZSL) learns $f_{CZSL}:\mathcal{X}\rightarrow\mathcal{Y}^{u}$, while generalized ZSL (GZSL) learns $f_{GZSL}:\mathcal{X}\rightarrow\mathcal{Y}^{s}\cup\mathcal{Y}^{u}$.

\subsection{Adaptive Attribute Distribution}
\textbf{Motivation. }
As illustrated in Fig.~\ref{fig:motivation}(a), static class-level semantic conditions cannot capture intra-class variations, leading to inaccurate visual–semantic correspondence and limited diversity of generated features. 
Although visual-guided semantic enhancement introduces instance-level semantics during training, it's difficult to be transferred to unseen classes due to the absence of visual features \cite{DSP23,VADS24}. 
Meanwhile, different attributes contribute differently to class discrimination (Fig.~\ref{fig:model_pipeline}(a)), motivating us to learn attribute-adaptive variation patterns. Therefore, we propose \textbf{Adaptive Attribute Distribution (AAD)}.

\textbf{Adaptive Attribute Distribution. }
Since intra-class visual variations usually occur around a class prototype, we aim to learn semantic variation patterns around the semantic prototype.
As shown in Fig.~\ref{fig:model_pipeline}(a), given a class-level attribute $a_c$, AAD learns latent attribute variation patterns for each dimension. Specifically, we introduce an Adaptive Attribute Encoder $\text{E}^{+}$, which predicts Gaussian distribution parameters from  $a_c=[a_c^1,\cdots,a_c^{|A|}] \in \mathbb{R}^{|A|}$:
\begin{equation}
   \label{E+}
   \text{E}^{+}(a_c) =\{(\mu_c^j,\sigma_c^j)\}_{j=1}^{|A|},
\end{equation}
where $\mu_c^j$ and $\sigma_c^j$ denote the mean and standard deviation of the $j$-th attribute dimension. Then the attribute variation is sampled as: $\Delta a_c^j\sim\mathcal{N}(\mu_c^j,(\sigma_c^j)^2),$ 
where $\Delta a_c^j$ represents the variation in the $j$-th attribute dimension. By using the $a_c$ as a semantic anchor, we obtain the diverse attribute:
\begin{equation}
    \label{eq:learned_variation}
    \tilde{a}^j_c = a_c^j + \lambda \cdot \Delta a_c^j,
\end{equation} where $\lambda$ controls the magnitude of the learned variation. Since instance-level semantic supervision is unavailable, uncontrollable variations may introduce unstable distributions during training. Therefore, we introduce a semantic consistency constraint $\mathcal{L}_{SC}$ to regularize the learned distributions while preserving semantic consistency:
$$ \mathcal{L}_{SC} = D_{KL} (\mathcal{N}(\mu_c^j,(\sigma_c^j)^2) \| \mathcal{N}(0,1)). $$

\textbf{Discriminability-guided Variation Calibration. }
Learned Attribute variations may weaken discriminative information. To balance diversity and discrimination, AAD calibrates the variation range according to attribute discriminability. We assume that highly discriminative attributes should vary less, while less discriminative attributes can accommodate larger variations. For the $j$-th attribute dimension, its discriminability is measured as:
\begin{equation}
   \label{d_j}
d_j=\frac{1}{|\mathcal{Y}^{s}|(|\mathcal{Y}^{s}|-1)}
\sum_{c\neq k}
(a_c^j-a_k^j)^2.
\end{equation}
A larger $d_j$ indicates stronger class-discriminative ability. After normalization, the standard deviation is calibrated as:
\begin{equation}
\label{eq:tilde_sigma}
\tilde{\sigma}_c^j=(1-\bar d_j)\sigma_c^j,
\end{equation}
where $\bar d_j$ denotes the normalized discriminability. With the calibrated standard deviation, the final attribute variation is sampled as: $\Delta \tilde{a}_c^j\sim\mathcal{N}(\mu_c^j,(\tilde{\sigma}_c^j)^2)$, and the final diverse attribute is obtained as $\tilde{a}_c^j = a_c^j + \lambda \cdot \Delta \tilde{a}_c^j$. Since attribute variations capture class-shared variation patterns, they can be transferred to unseen classes.

\subsection{Visual Structure Alignment}
As illustrated in Fig.~\ref{fig:motivation}(b), structurally misaligned semantic conditions may cause the synthesized features to deviate from the real visual distribution. 
To address this issue, we propose \textbf{Visual Structure Alignment (VSA)} to learn semantic representations with visual-consistent inter-class structures, as shown in Fig.~\ref{fig:model_pipeline}(b).
During training, we use a frozen feature encoder $\text{F}$ to extract visual features from seen-class images. 
Since pre-trained visual representations may contain class-shared common directions, we first compute class prototypes: $p_c=\frac{1}{N_c}\sum_{i:y_i=c}x_i,$ and then suppress the dominant common direction identified by SVD:
\begin{equation}
\label{eq:p_hat}
\hat{p}_c
=
p_c-\alpha(p_c^\top p_{\mathrm{com}})p_{\mathrm{com}},
\end{equation}
where $p_{\mathrm{com}}$ is the unit-normalized leading right singular vector of the seen-class prototype matrix and $\alpha$ controls the suppression strength. The refined prototypes enhance inter-class discriminability while preserving class structures. 
Based on \(\tilde{a}_c\) learned by AAD, we introduce a Structure Alignment Encoder \(\text{E}^*\) to obtain attribute embeddings: $v_c = \text{E}^* (\tilde{a}_c)$, optimized by the prototype alignment loss $\mathcal{L}_{PA}$:
\begin{equation}
   \label{L_PA}
 \mathcal{L}_{PA} = -\log \frac{ \exp(v_c^\top \hat{p}_c/\tau) } { \sum_{k\in\mathcal{Y}^{s}} \exp(v_c^\top \hat{p}_k/\tau) }.
\end{equation}
This prototype-guided alignment enables \(v_c\) to capture discriminative visual structures while maintaining the semantic transferability of attribute-based conditions. 

\subsection{Optimization and Unseen-class Feature Generation}
\textbf{Objective Loss Function. } 
Overall, the total loss function is defined as follows:
\begin{equation}
    \label{L_total}
\mathcal{L}_{total}=\mathcal{L}_{G}+\lambda_{SC}\mathcal{L}_{SC}+\lambda_{PA}\mathcal{L}_{PA},
\end{equation}
where $\mathcal{L}_{G}$ is the loss of the conditional generator $\text{G}$. $\lambda_{SC}$ and $\lambda_{PA}$ are the weights for $\mathcal{L}_{SC}$ and $\mathcal{L}_{PA}$. After training, we use the learned generator $\text{G}$ to synthesize unseen-class visual features for ZSL classification. We also provide a detailed training algorithm in Appendix~\ref{app:full_algorithm}.

\textbf{Unseen Class Feature Generation. } 
As shown by the \textcolor{mygreen}{green arrows} in Fig.~\ref{fig:model_pipeline}, given the class-level attribute \(a_u\) of an unseen class \(u\), the trained \(\text{E}^{+}\) first constructs adaptive attribute distributions through AAD, i.e., \(\tilde{a}_u^j \sim \mathcal{N}(a_u^j+\lambda \mu_u^j,(\lambda\tilde{\sigma}_u^j)^2)\). The sampled attributes \(\tilde{a}_u\) are then projected by \(\text{E}^*\) to obtain attribute embedding, i.e., \(v_u=\text{E}^*(\tilde{a}_u)\). Finally, as shown in Fig.~\ref{fig:model_pipeline}(c), \(\tilde{a}_u\) and \(v_u\) are concatenated as the generation condition, which is combined with sampled latent noise $z$ to synthesize unseen-class visual features, i.e., \(\hat{x}_u=\text{G}(z; [\tilde{a}_u,v_u])\). Detailed evaluation procedures are in Appendix~\ref{app:full_algorithm}.

\section{Experiments}
\label{Experiments}

\begin{table*}[t]
    \centering
    \small
    \setlength{\tabcolsep}{3.2pt}
    \renewcommand{\arraystretch}{1.08}
    \caption{
        Comparison with state-of-the-art methods on AWA2, CUB, and SUN.
        Acc, U, S, and H (\%) are reported.
        Avg. H averages H across all three datasets when available.
        $\dagger$ indicates methods using class names as semantic conditions.
        Best and second-best results are shown in \textbf{bold} and \underline{underlined}, respectively and -- denotes unreported results.
    }
    \label{tab:main_results}
    \resizebox{\textwidth}{!}{%
    \begin{tabular}{l l *{13}{c}}
        \toprule
        \multirow{2}{*}{\textbf{Method}} &
        \multirow{2}{*}{\textbf{Venue}} &
        \multicolumn{4}{c}{\textbf{AWA2}} &
        \multicolumn{4}{c}{\textbf{CUB}} &
        \multicolumn{4}{c}{\textbf{SUN}} &
        \multirow{2}{*}{\textbf{Avg. H}} \\
        \cmidrule(lr){3-6}
        \cmidrule(lr){7-10}
        \cmidrule(lr){11-14}
        & &
        \textbf{Acc} & \textbf{U} & \textbf{S} & \textbf{H} &
        \textbf{Acc} & \textbf{U} & \textbf{S} & \textbf{H} &
        \textbf{Acc} & \textbf{U} & \textbf{S} & \textbf{H} & \\
        \midrule

        \multicolumn{15}{l}{\textit{Embedding-based}} \\
        \addlinespace[1pt]

        CLIP$^{\dagger}$
        & ICML'21
        & -- & -- & -- & --
        & -- & 55.2 & 54.8 & 55.0
        & -- & -- & -- & --
        & -- \\

        PSVMA
        & CVPR'23
        & -- & 73.6 & 77.3 & 75.4
        & -- & 70.1 & 77.8 & 73.8
        & -- & 61.7 & 45.3 & 52.3
        & 67.2 \\

        ZSLViT
        & CVPR'24
        & 70.2 & 66.1 & 84.6 & 74.2
        & 78.9 & 69.4 & 78.2 & 73.6
        & 68.3 & 45.9 & 48.3 & 47.3
        & 65.0 \\

        ZeroMamba
        & AAAI'25
        & 71.9 & 67.9 & 87.6 & 76.5
        & 80.0 & 72.1 & 76.4 & 74.2
        & 72.4 & 56.5 & 41.4 & 47.7
        & 66.1 \\

        CSAP-ZSL
        & AAAI'25
        & -- & 66.8 & 84.4 & 74.6
        & -- & 68.1 & 73.2 & 70.6
        & -- & 61.3 & 45.5 & 52.2
        & 65.8 \\

        AENet
        & AAAI'25
        & 75.2 & 70.3 & 80.1 & 74.9
        & 80.3 & 73.1 & 76.4 & 74.7
        & 70.4 & 58.6 & 45.2 & 51.0
        & 66.9 \\

        VSPCN
        & CVPR'25
        & 76.6 & 71.8 & 84.3 & 77.6
        & 80.6 & 72.8 & 78.9 & 75.7
        & 75.3 & 59.4 & 49.1 & 53.8
        & 69.0 \\

        SVIP
        & ICCV'25
        & 69.8 & 65.4 & 87.7 & 74.9
        & 79.8 & 72.1 & 78.1 & 75.0
        & 71.6 & 53.7 & 48.0 & 50.7
        & 66.9 \\

        PSVMA+
        & TPAMI'25
        & 79.2 & 74.2 & 86.4 & 79.8
        & 78.8 & 71.8 & 77.8 & 74.6
        & 74.5 & 61.5 & 49.4 & 54.8
        & 69.7 \\

        SAGE
        & AAAI'26
        & \underline{83.7} & \underline{78.1} & \underline{87.2} & \underline{82.4}
        & \underline{82.1} & \textbf{75.1} & 77.5 & \underline{76.3}
        & \textbf{77.2} & \underline{63.7} & 52.2 & \textbf{57.4}
        & \underline{72.0} \\

        \addlinespace[2pt]
        \multicolumn{15}{l}{\textit{Generative-based}} \\
        \addlinespace[1pt]

        f-VAEGAN
        & CVPR'19
        & 71.1 & 57.6 & 70.6 & 63.5
        & 61.0 & 48.4 & 60.1 & 53.6
        & 64.7 & 45.1 & 38.0 & 41.3
        & 52.8 \\

        TF-VAEGAN
        & ECCV'20
        & 72.2 & 59.8 & 75.1 & 66.6
        & 64.9 & 52.8 & 64.7 & 58.1
        & 66.0 & 45.6 & 40.7 & 43.0
        & 55.9 \\

        ICCE
        & CVPR'22
        & 78.4 & 65.3 & 82.3 & 72.8
        & 72.7 & 67.3 & 65.5 & 66.4
        & -- & -- & -- & --
        & -- \\

        ZeroNAS
        & TPAMI'22
        & 73.2 & 61.4 & 75.3 & 67.6
        & 66.4 & 56.0 & 63.8 & 59.6
        & 68.3 & 47.1 & 41.8 & 44.3
        & 57.2 \\

        CDL + OSCO
        & TPAMI'23
        & -- & 48.0 & 71.0 & 57.1
        & -- & -- & -- & --
        & -- & 32.0 & \textbf{65.0} & 42.9
        & -- \\

        DSP
        & ICML'23
        & 71.6 & 60.0 & 86.0 & 70.7
        & 62.8 & 51.4 & 63.8 & 56.9
        & 68.6 & 48.3 & 43.0 & 45.5
        & 57.7 \\

        TF-VAEGAN + SHIP$^{\dagger}$
        & ICCV'23
        & -- & 61.2 & \textbf{95.9} & 74.7
        & -- & 22.5 & \textbf{82.2} & 35.3
        & -- & -- & -- & --
        & -- \\

        DML
        & ACCV'24
        & -- & 62.2 & 82.3 & 70.9
        & -- & 57.1 & \underline{81.6} & 67.2
        & -- & 39.6 & 52.7 & 45.9
        & 61.3 \\

        VADS
        & CVPR'24
        & 82.5 & 75.4 & 83.6 & 79.3
        & \textbf{86.8} & 74.1 & 74.6 & 74.3
        & \underline{76.3} & \textbf{64.6} & 49.0 & 55.7
        & 69.8 \\

        ViFR
        & IJCV'25
        & 77.8 & 68.2 & 78.9 & 73.2
        & 74.5 & 63.9 & 72.0 & 67.7
        & 69.2 & 51.3 & 40.0 & 44.7
        & 61.9 \\

        \rowcolor{gray!12}
        \textbf{AAVS (Ours)}
        & --
        & \textbf{85.4} & \textbf{82.1} & 85.0 & \textbf{83.5}
        & 79.2 & \underline{74.2} & 81.0 & \textbf{77.4}
        & 75.8 & 58.9 & \underline{54.0} & \underline{56.3}
        & \textbf{72.4} \\

        \bottomrule
    \end{tabular}
    }
    \setlength{\textfloatsep}{-2pt}
\end{table*}

\subsection{Datasets and Evaluation Protocols}
We evaluate AAVS on three ZSL benchmarks: AWA2 \citep{ZSL19}, CUB \citep{CUB}, and SUN \citep{SUN}. The statistics of these datasets, including the numbers of images, classes, and attributes, are $\{37322,50,85\}$, $\{11788,200,312\}$, and $\{14340,717,102\}$, respectively. We follow the standard class split protocol \citep{ZSL19}. Details are in Appendix~\ref{app:dataset}. For evaluation, we adopt the unified protocol \citep{ZSL19}. We report unseen-class accuracy ($\text{Acc}$) in CZSL, and unseen ($\text{U}$), seen ($\text{S}$), and harmonic mean $\text{H}=\frac{2\times\text{S}\times\text{U}}{\text{S}+\text{U}}$ in GZSL.

\subsection{Implementation Details}
We adopt f-VAEGAN \citep{f-VAEGAN19} as the conditional generator to synthesize visual features, which combines VAE and GAN for semantic-conditioned generation. 
Visual features are extracted from a pre-trained ViT \citep{DINOv2} using the CLS token. For the fine-grained CUB dataset, we further fine-tune ViT to alleviate dataset bias \citep{FREE21,ViFR25}. 
For the semantic consistency constraint, we set its weight $\lambda_{SC}$ to 0.01 for all datasets to balance distribution learning and semantic consistency. 
The magnitude of learned variation $\lambda$ (Eq.~\ref{eq:learned_variation}), the strength of common direction suppression are controlled by the parameter $\alpha$ (Eq. \ref{eq:p_hat}) and the number of synthesized features per class for AWA2, CUB, and SUN are $\{0.05,0.1,0.01\}$, $\{0.5,0.5,0.5\}$, and $\{1800,500,400\}$, respectively. 
All experiments are implemented in PyTorch and conducted on a single NVIDIA RTX 3090 GPU (24 GB).

\subsection{Comparison with State-of-the-Art Methods}
As shown in Table~\ref{tab:main_results}, we compare AAVS, built upon f-VAEGAN \citep{f-VAEGAN19}, with state-of-the-art embedding-based and generative ZSL methods.  
Under CZSL, AAVS achieves 85.4\% accuracy on AWA2, outperforming SAGE \citep{SAGE26} (83.7\%) by 1.7\%. It also obtains 79.2\% and 75.8\% on CUB and SUN, respectively, demonstrating competitive performance among generative methods.
Under GZSL, AAVS achieves the highest average H of 72.4\% among all compared embedding-based and generative methods. This verifies that AAD and VSA provide effective semantic conditions by capturing attribute variations and visual-consistent structures.
Furthermore, AAVS remains competitive with recent ViT-based methods, including SVIP \citep{SVIP25}, PSVMA+ \citep{PSVMA+25}, and SAGE \citep{SAGE26}. It also surpasses DSP \citep{DSP23} and VADS \citep{VADS24}, which rely on visual-guided semantic enhancement, highlighting the advantage of learning transferable attribute distributions without relying on visual features. 
VLM-based methods such as SHIP \citep{SHIP23} achieve strong seen-class performance but exhibit limited generalization to unseen classes.

\subsection{Ablation Study}
To evaluate the contribution of each component in AAVS, we remove Adaptive Attribute Distribution (AAD), Visual Structure Alignment (VSA), and their key designs on AWA2, CUB, and SUN, as shown in Table~\ref{tab:ablation}. \textbf{(a)} Removing both AAD and VSA reduces AAVS to a baseline that directly uses class-level attributes for generation.
Compared with the baseline, AAVS consistently improves performance across all datasets, e.g., increasing H from 65.4\% to 83.5\% on AWA2 and by 5.1\% on CUB. This demonstrates that jointly modeling adaptive attribute distributions and visual structure alignment provides more effective generation conditions than static class-level semantics.
\textbf{(b)} Removing AAD degrades performance by 11.0\%, 4.3\%, and 0.5\% in H on AWA2, CUB, and SUN, respectively. This is because the model relies on static class-level semantics and lacks the ability to capture intra-class variations, highlighting the importance of semantic diversity. 
\textbf{(c)} Removing VSA leads to H decreases of 8.2\%, 1.0\%, and 2.5\%, validating the importance of visual structure consistency for generating features aligned with real distributions.
\textbf{(d, e)} Furthermore, removing either Eq.~(\ref{eq:tilde_sigma}) or Eq.~(\ref{eq:p_hat}) consistently reduces performance, validating that variation calibration balances semantic diversity and class discrimination, while prototype refinement provides more discriminative supervision for visual structure alignment.

\begin{table*}[t]
    \centering
    \caption{
        Ablation results of AAVS on AWA2, CUB, and SUN.
        AAVS w/o Eq.~(\ref{eq:tilde_sigma}) ablates discriminability-guided
        variation calibration, while AAVS w/o Eq.~(\ref{eq:p_hat}) ablates
        visual prototype refinement.
        Avg. H averages H across all three datasets.
    }
    \label{tab:ablation}
    \small
    \setlength{\tabcolsep}{2.5pt}
    \renewcommand{\arraystretch}{1.08}
    \begin{tabular*}{\textwidth}{
        @{\extracolsep{\fill}} l *{13}{c} @{}
    }
        \toprule
        \multirow{2}{*}{\textbf{Configuration}}
        & \multicolumn{4}{c}{\textbf{AWA2}}
        & \multicolumn{4}{c}{\textbf{CUB}}
        & \multicolumn{4}{c}{\textbf{SUN}}
        & \multirow{2}{*}{\textbf{Avg. H}} \\
        \cmidrule(lr){2-5}
        \cmidrule(lr){6-9}
        \cmidrule(lr){10-13}
        & \textbf{Acc} & \textbf{U} & \textbf{S} & \textbf{H}
        & \textbf{Acc} & \textbf{U} & \textbf{S} & \textbf{H}
        & \textbf{Acc} & \textbf{U} & \textbf{S} & \textbf{H}
        & \\
        \midrule

        a) AAVS w/o AAD \& VSA
        & 64.1 & 62.4 & 68.7 & 65.4
        & 77.1 & 66.8 & 78.7 & 72.3
        & 73.7 & 57.4 & 50.0 & 53.5
        & 63.7 \\

        b) AAVS w/o AAD
        & 71.2 & 63.6 & 84.4 & 72.5
        & 76.5 & 68.0 & 78.9 & 73.1
        & 74.2 & \textbf{59.9} & 52.1 & 55.8
        & 67.1 \\

        c) AAVS w/o VSA
        & 78.8 & 70.6 & 80.6 & 75.3
        & \textbf{79.2} & 73.2 & \underline{79.9} & 76.4
        & 73.9 & 57.4 & 50.6 & 53.8
        & 68.5 \\

        d) AAVS w/o Eq.~(\ref{eq:tilde_sigma})
        & 84.8 & \underline{78.8} & \underline{85.6} & 81.9
        & 78.6 & 72.7 & 79.2 & 76.0
        & 74.2 & 58.8 & \underline{53.2} & \underline{55.9}
        & 71.3 \\

        e) AAVS w/o Eq.~(\ref{eq:p_hat})
        & \underline{85.2} & 78.2 & \textbf{86.6} & \underline{82.3}
        & \underline{78.9} & \underline{74.1} & 79.8 & \underline{76.8}
        & \underline{75.3} & \underline{59.7} & 52.3 & 55.8
        & \underline{71.6} \\

        \midrule
        f) AAVS
        & \textbf{85.4} & \textbf{82.1} & 85.0 & \textbf{83.5}
        & \textbf{79.2} & \textbf{74.2} & \textbf{81.0} & \textbf{77.4}
        & \textbf{75.8} & 58.9 & \textbf{54.0} & \textbf{56.3}
        & \textbf{72.4} \\
        \bottomrule
    \end{tabular*}
    \setlength{\textfloatsep}{0pt}
\end{table*}

\subsection{Qualitative Evaluation}
\subsubsection{Analysis of Learned Generation Conditions}
\begin{figure*}[t]
    \centering
    \includegraphics[width=\textwidth]{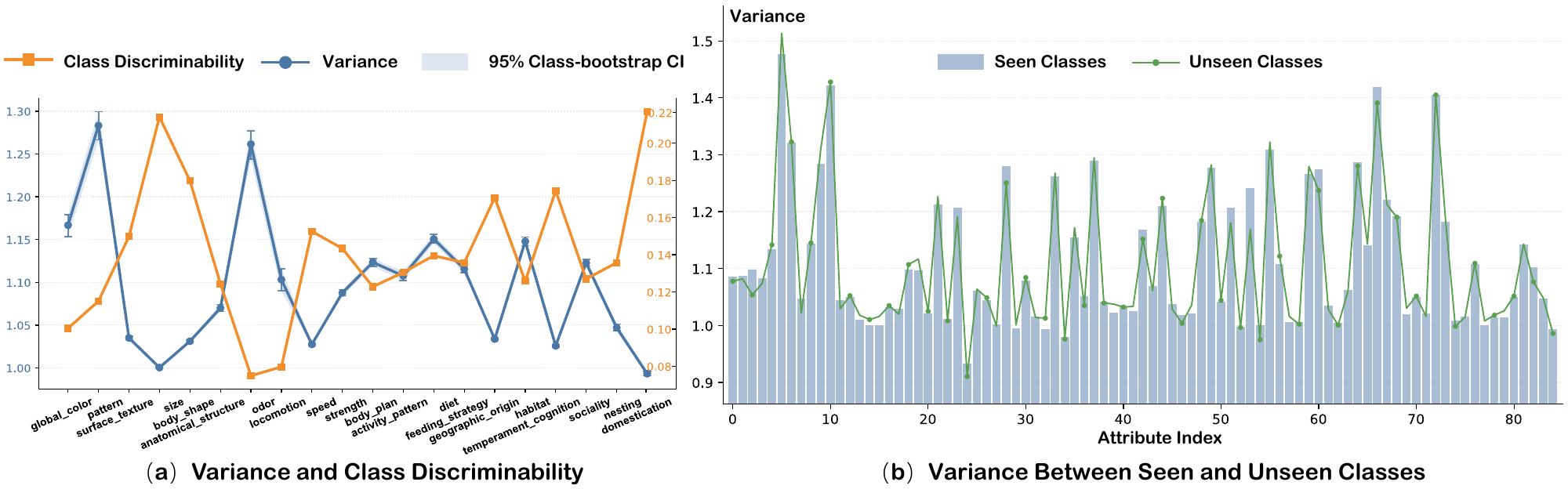}
    \setlength{\abovecaptionskip}{0pt}
    \setlength{\belowcaptionskip}{-2pt}
    \caption{ 
    Qualitative analysis of the Adaptive Attribute Distribution (AAD) on AWA2.
    (a) Attribute discriminability versus learned variance $\sigma$: more discriminative attributes receive smaller variations, while less discriminative attributes allow larger variations.
(b) Seen--unseen comparison of attribute-wise variation patterns, demonstrating that AAD transfers through the shared attribute space without unseen-class visual features.
    }
    \label{fig:proof_for_AAD}
\end{figure*}

\textbf{Semantic Interpretability of Adaptive Attribute Distributions. }
To verify whether the learned attribute Distributions capture meaningful semantic patterns, we analyze their relationship with attribute inter-class discriminability (Eq.~\ref{d_j}) on AWA2. 
We group attributes by semantic types (e.g., Black and White as global color) and compute the average variance and discriminability within each group, as shown in Fig.~\ref{fig:proof_for_AAD}(a).
The results reveal a negative correlation between discriminability and learned variation magnitude: highly discriminative attributes (e.g., size) exhibit smaller variations, while less discriminative attributes (e.g., pattern) allow larger variations, consistent with our motivation. This indicates that AAD adaptively allocates variation capacity across attributes, preserving discriminative semantics while introducing additional diversity beyond the class prototype.

\textbf{Transferability of Learned Attribute Variations.}
For generative ZSL, transferring diverse semantic conditions to unseen classes is essential. Fig.~\ref{fig:proof_for_AAD}(b) compares attribute variation patterns between seen and unseen classes on AWA2.
Although the variation patterns are learned only from seen classes, they exhibit similar attribute-level trends when applied to unseen classes. 
This demonstrates that AAD can transfer learned variation patterns through the shared attribute space, enabling unseen classes to obtain class-adaptive diverse attributes without visual observations or instance-level semantic annotations. 

\begin{figure*}[t]
    \centering
    \includegraphics[width=\textwidth]{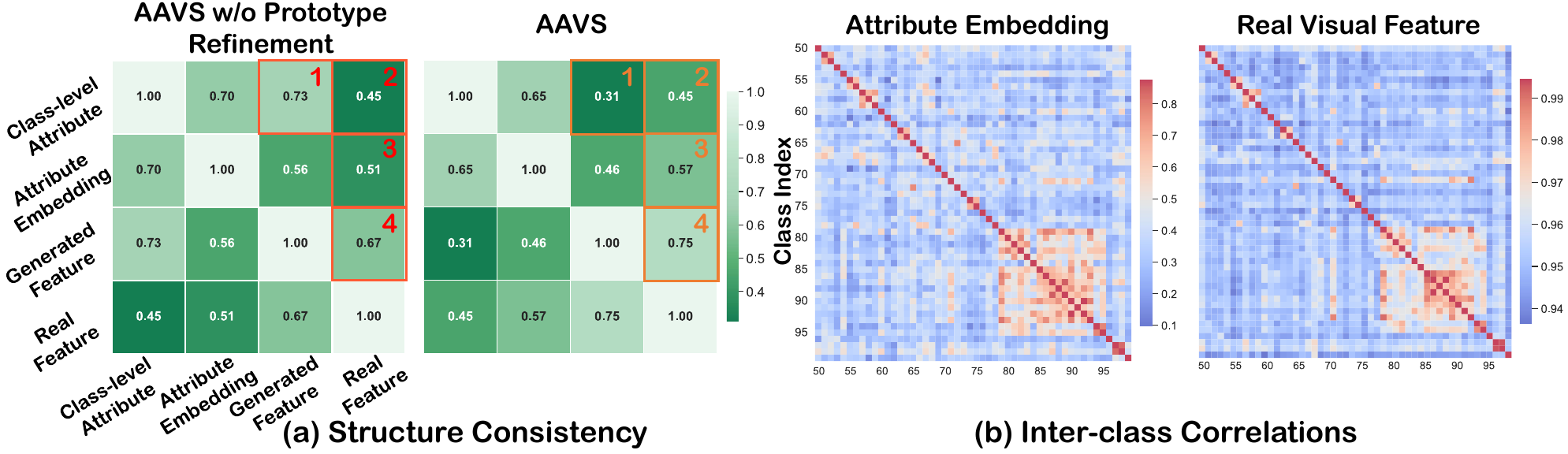}
    \setlength{\abovecaptionskip}{0pt}
    \setlength{\belowcaptionskip}{-2pt}
    \caption{ 
Qualitative evaluation of VSA on CUB.
(a) Spearman's $\rho$ \citep{spearman1961proof} matrices for four representations with and without prototype refinement; refinement increases the correlation between attribute embeddings and real visual features from $0.51$ to $0.57$ compared to class-level attributes (\textcolor{myred}{box 3} to \textcolor{myorange}{box 3}).
(b) Inter-class correlation matrix of the attribute embeddings and real visual features, which also shows the effectiveness of VSA.}
    \label{fig:proof_for_VSA}
\end{figure*}

\textbf{Visual Structure Consistency of Semantic Embeddings. }
Fig.~\ref{fig:proof_for_VSA} evaluates the effectiveness of VSA on the CUB dataset from two perspectives. 
As shown in Fig.~\ref{fig:proof_for_VSA}(a), generated features largely inherit the inter-class structure of class attributes (\textcolor{myred}{box 1}, Spearman's $\rho$ of 0.73), validating our motivation. 
In contrast, VSA-learned attribute embeddings $v$ exhibit stronger structural consistency with real visual features (\textcolor{myred}{box 2} and \textcolor{myred}{box 3}, 0.45 vs. 0.51), indicating more visually aligned semantic conditions.
Prototype refinement in Eq.~(\ref{eq:p_hat}) further alleviates similarity compression in pre-trained visual representations (details in Appendix~\ref{app:prototype_refinement}). 
After refinement, the inherited structural discrepancy is reduced from 0.73 to 0.31 (\textcolor{myred}{box 1} and \textcolor{myorange}{box 1}), while the alignment between attribute embeddings $v$ and visual structures is improved (\textcolor{myorange}{box 3}, 0.57).
The correlation matrices in Fig.~\ref{fig:proof_for_VSA}(b) further visualize this enhanced alignment, demonstrating the effectiveness of VSA.

\subsubsection{Quality of Generated Features}
To qualitatively assess the generated features, we use the t-SNE \citep{tsne2008} to visualize them produced by f-VAEGAN \citep{f-VAEGAN19}, f-VAEGAN with AAD, f-VAEGAN with VSA, and the full AAVS model in Fig.~\ref{fig:TSNE_CUB}.
As detailed in Appendix~\ref{app:feature}, the baseline generates overly compact intra-class distributions and exhibits a mismatch with the real inter-class structure. AAD alleviates the former by introducing adaptive attribute variations, while VSA addresses the latter through visually aligned attribute embeddings, as further illustrated by \textcolor{myorange}{box 4} in Fig.~\ref{fig:proof_for_VSA}(a). Combining both components, AAVS produces features with greater intra-class diversity and better consistency with the real visual structure, in line with our motivation.

\subsection{Hyper-Parameter Analysis}
\begin{figure*}[t]
    \centering
    \includegraphics[width=\textwidth]{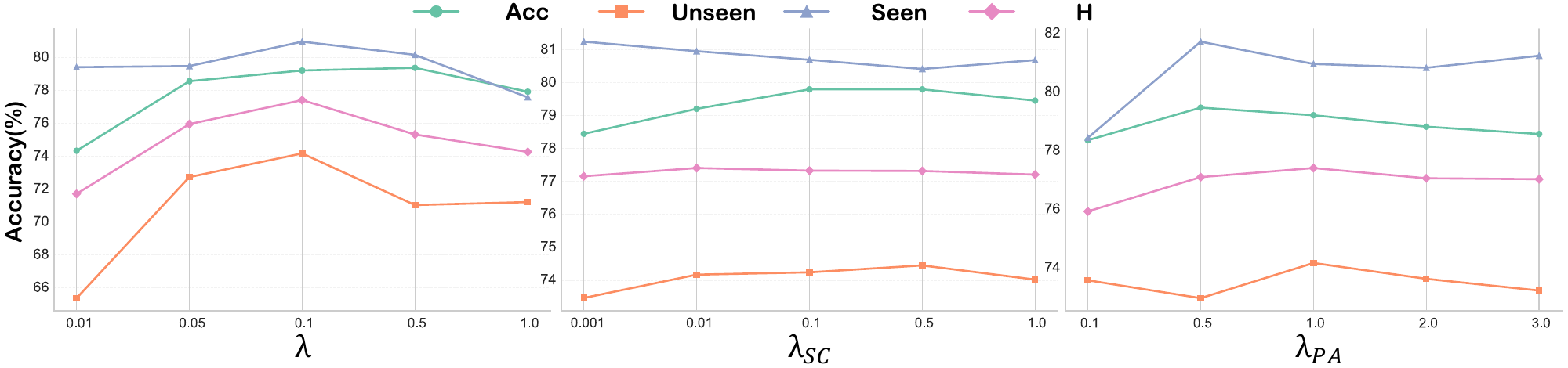}
    \setlength{\abovecaptionskip}{0pt}
    \setlength{\belowcaptionskip}{-2pt}
    \caption{ 
    Hyper-parameter analysis on CUB. $\lambda$ denotes the magnitude of learned variation in Eq.~(\ref{eq:learned_variation}), $\lambda_{SC}$ and $\lambda_{PA}$ control the weights of two losses in Eq.~(\ref{L_total}), respectively.
    }
    \label{fig:ablation_on_CUB}
\end{figure*}

Fig.~\ref{fig:ablation_on_CUB} analyzes the sensitivity of $\lambda$, $\lambda_{SC}$, and $\lambda_{PA}$ on CUB. 
As $\lambda$ controls the magnitude of semantic variations in AAD, it determines the trade-off between diversity and consistency. 
Too small variations limit semantic diversity, while excessive variations deviate from semantic prototypes and harm performance, demonstrating the importance of proper variation control. 
In contrast, $\lambda_{SC}$ and $\lambda_{PA}$ remain stable across different values. Results on AWA2 and SUN are reported in Appendix~\ref{app:hyper}.

\section{Conclusion}
In this paper, we analyse the limitations of class-level semantic conditions in generative ZSL and identify two essential properties of effective generation conditions: semantic diversity and visual-structure consistency.
Based on these properties, we proposed AAVS, in which AAD learns transferable, adaptive attribute distributions to produce diverse attributes, while VSA aligns attribute embeddings with visual class space to learn inter-class structure.
Together, they construct effective semantic conditions that transfer from seen to unseen classes and guide the synthesis of higher-quality visual features. Extensive experiments on three benchmarks demonstrate that AAVS outperforms existing methods and validate the effectiveness of learned semantic conditions.

\subsection*{AI use statement}
In this work, we used generative AI tools for language editing and drafting portions of the manuscript. Generative AI was not used to generate or modify experimental data, conduct experiments, or create any figures or tables in the manuscript. All AI-assisted content was reviewed and revised by the authors. We take full responsibility for the final content of this work, including all text, claims, and materials produced with the assistance of generative AI.

\subsection*{Ethics statement}
This study uses publicly available ZSL benchmark datasets and does not involve human participants, private information, or newly collected data.

\subsection*{Reproducibility statement}

To facilitate reproducibility, we provide a code repository in the supplementary material and dataset splits in the appendix, while all hyperparameters are specified in the main text. The repository includes features extraction scripts for all benchmark datasets, complete training and evaluation code, and detailed instructions for running the main experiments. All reported experiments can be conducted on a single NVIDIA RTX 3090 GPU.



\bibliography{iclr2027_conference}
\bibliographystyle{iclr2027_conference}

\newpage
\appendix
\section{Appendix}
\subsection{Prototype Refinement Analysis}
\label{app:prototype_refinement}
Prototype refinement in Eq.~(\ref{eq:p_hat}) aims to alleviate similarity compression while preserving the real inter-class structure. We investigate the effect of the suppression strength \(\alpha\) on refined prototypes and downstream ZSL performance.

\begin{figure*}[b]
    \centering
    \includegraphics[width=\textwidth]{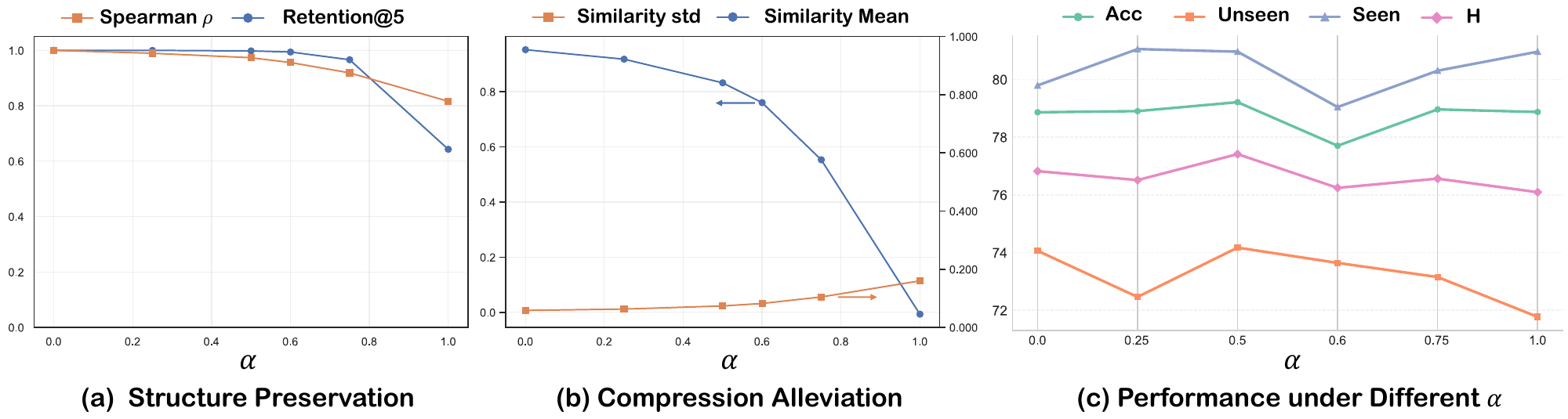}
    \caption{
Analysis of the prototype refinement strength $\alpha$ on CUB.
(a) Refinement preserves the original prototype structure measured by Spearman $\rho$ and Retention@5.
(b) Refinement alleviates similarity compression by reducing the mean similarity and increasing the similarity variance.
(c) Moderate refinement improves ZSL performance, while excessive refinement causes structural deviation and degrades unseen-class generalization.
}
    \label{fig:proof_for_alpha}
\end{figure*}

A larger \(\alpha\) removes more shared representation, improving class separability but potentially distorting the real structure. As shown in Fig.~\ref{fig:proof_for_alpha}(a), increasing \(\alpha\) gradually decreases Spearman $\rho$ and Retention@5, indicating increasing structural distortion. However, both metrics remain stable under moderate refinement, demonstrating that the refined prototypes learn the real class relationships.

Fig.~\ref{fig:proof_for_alpha}(b) shows that increasing \(\alpha\) reduces the mean similarity while increasing its standard deviation, indicating that refinement expands the dynamic range of visual similarity and alleviates similarity compression.

Fig.~\ref{fig:proof_for_alpha}(c) further shows the impact of \(\alpha\) on ZSL performance. Prototype refinement improves performance, validating its effectiveness. However, excessive refinement improves seen accuracy while degrading unseen accuracy, as over-suppression may distort visual class structures and lead VSA to learn biased structures that fit seen classes but fail to generalize to unseen classes. Therefore, we set \(\alpha=0.5\) to balance structural preservation and compression alleviation.

\subsection{Hyper-parameter Analysis on AWA2 and SUN}
\label{app:hyper}
\begin{figure*}[t]
    \centering
    \includegraphics[width=\textwidth]{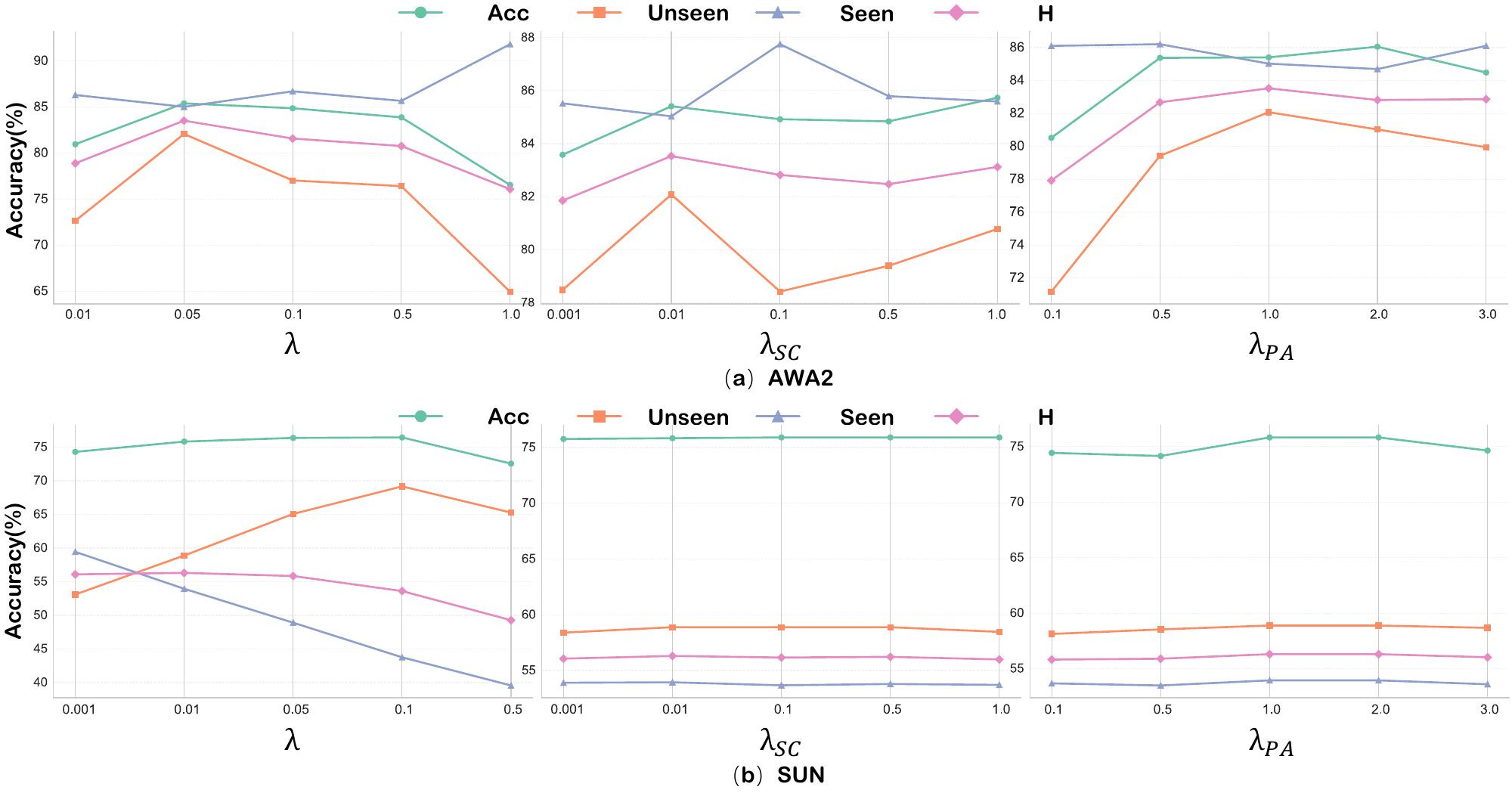}
    \caption{ 
    Hyper-parameter analysis on AWA2 and SUN. $\lambda$ denotes the magnitude of learned variation in Eq.~(\ref{eq:learned_variation}), $\lambda_{SC}$ and $\lambda_{PA}$ control the weights of two losses in Eq.~(\ref{L_total}), respectively.
    }
    \label{fig:ablation_on_AWA2_SUN}
\end{figure*}

We further provide sensitivity analyses of $\lambda$, $\lambda_{SC}$, and $\lambda_{PA}$ on AWA2 and SUN in Fig.~\ref{fig:ablation_on_AWA2_SUN}. 
Among them, $\lambda$ controls the magnitude of AAD-induced semantic variations and directly affects the trade-off between semantic diversity and consistency. 
A small $\lambda$ limits the utilization of learned attribute variations, resulting in insufficiently diverse semantic conditions and degraded unseen-class generation. 
In contrast, an excessively large $\lambda$ introduces variations beyond the semantic prototype, causing performance degradation on both seen and unseen classes. It may even lead to overfitting to seen classes, as observed on AWA2 when $\lambda=1.0$. 
These results further highlight the importance of properly controlling variation magnitude for effective semantic condition learning. 
Meanwhile, $\lambda_{SC}$ and $\lambda_{PA}$ remain stable over a wide range of values, demonstrating the robustness of our method.

\subsection{Qualitative Analysis of Generated Feature}
\label{app:feature}
To qualitatively analyze the impact of each AAVS component on generated feature quality, we visualize and compare the feature distributions of f-VAEGAN (baseline), AAVS, f-VAEGAN+AAD, and f-VAEGAN+VSA using t-SNE \citep{tsne2008}.

\begin{figure*}[t]
    \centering
    \includegraphics[width=\textwidth]{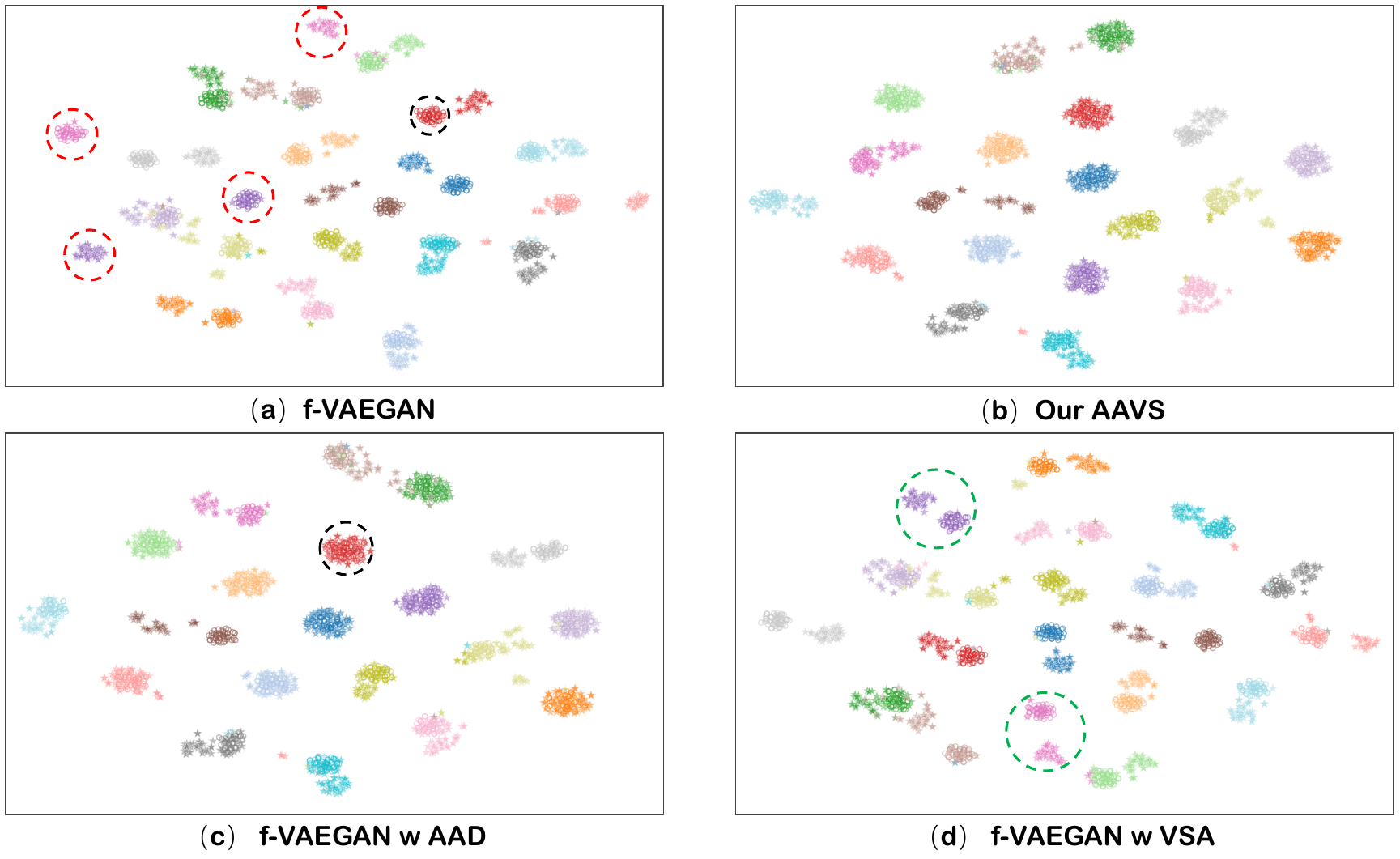}
    \caption{ 
    Qualitative evaluation of AAVS on CUB.
    We randomly select 10 seen and 10 unseen classes, with 50 generated and 50 real features sampled per class. Different colors indicate classes, while $\circ$ and $\star$ denote generated and real features, respectively. (Zoom in for details.) 
    }
    \label{fig:TSNE_CUB}
\end{figure*}

The baseline suffers from distribution shifts between generated and real features (\textcolor{myred}{red circles} in Fig.~\ref{fig:TSNE_CUB}(a)), as it inherits semantic structures inconsistent with the visual space. 
VSA mitigates this issue by introducing visually aligned attribute embeddings, producing features closer to the real distribution (\textcolor{mygreen}{green circles} in Fig.~\ref{fig:TSNE_CUB}(d)).

Moreover, static class-level attributes lead to concentrated generated features with limited intra-class diversity (\textbf{black circles} in Fig.~\ref{fig:TSNE_CUB}(a)), while AAD improves diversity by learning adaptive attribute distributions (\textbf{black circles} in Fig.~(c)). 

Combining both components, AAVS generates features with improved diversity and better visual distribution alignment, as shown in Fig.~\ref{fig:TSNE_CUB}(b).

\subsection{Benchmark Datasets}
\label{app:dataset}
We evaluate AAVS on three widely used ZSL benchmarks under the Proposed Split (PS) protocol \citep{ZSL19}, as summarized in Table~\ref{tab:dataset}.
AWA2 is a coarse-grained animal recognition dataset containing 37,322 images from 50 classes, each represented by an 85-dimensional attribute vector.
CUB is a fine-grained bird dataset with 11,788 images from 200 species and 312-dimensional attributes.
SUN is a large-scale scene recognition dataset comprising 14,340 images from 717 scene categories, described by 102-dimensional attributes.

Following PS, AWA2 is divided into 40 seen and 10 unseen classes, CUB into 150 seen and 50 unseen classes, and SUN into 645 seen and 72 unseen classes.
Only images from seen classes are available for model training, while unseen-class images are reserved for evaluation.
Under CZSL, classification is restricted to unseen classes; under GZSL, both seen and unseen classes constitute the test label space.
\begin{table}[t]
    \centering
    \caption{
        Dataset statistics under the Proposed Split (PS).
        $s/u$ denotes the numbers of seen/unseen classes, and $|A|$ is the attribute dimension.
    }
    \label{tab:dataset}
    \small
    \renewcommand{\arraystretch}{1.08}
    \begin{tabular*}{\columnwidth}{
        @{\extracolsep{\fill}} l c c c @{}
    }
        \toprule
        \textbf{Dataset} &
        \textbf{\#Images} &
        \textbf{\#Classes ($s/u$)} &
        \textbf{$|A|$} \\
        \midrule
        AWA2 \citep{ZSL19} & 37,322 & 50 (40/10)  & 85  \\
        CUB  \citep{CUB}   & 11,788 & 200 (150/50) & 312 \\
        SUN  \citep{SUN}   & 14,340 & 717 (645/72) & 102 \\
        \bottomrule
    \end{tabular*}
\end{table}

\subsection{Full algorithm}
\label{app:full_algorithm}
Algorithms~\ref{alg:train_aavs} and~\ref{alg:evaluate_zsl} summarize the training and evaluation procedures of AAVS.

During training, we construct refined prototypes and compute the inter-class discriminability. For each seen-class feature $x_i^s$, $\mathrm{E}^{+}$ predicts an adaptive attribute distribution and samples $\tilde{a}_i$, while $\mathrm{E}^{*}$ maps it to $v_i$. Their concatenation forms the generation condition, and $\mathrm{E}^{+}$, $\mathrm{E}^{*}$, and the generator $G$ are jointly optimized.

At inference, AAVS constructs conditions solely from unseen-class attributes and synthesizes $N_{\mathrm{syn}}$ features per unseen class using distinct semantic conditions. We then train the corresponding classifier and evaluate it on the test data under each ZSL setting.

\begin{algorithm*}[t]
\caption{Training AAVS and the Conditional Generator}
\label{alg:train_aavs}
\begin{algorithmic}
\Require Seen-class training data $\mathcal{D}_{\mathrm{tr}}^{s}$,
class attributes $\{a_c\}_{c\in\mathcal{Y}^{s}}$,
visual encoder $F$, number of training epochs $T$,
and hyperparameters $\lambda$, $\lambda_{SC}$, and $\lambda_{PA}$
\Ensure Trained adaptive attribute encoder $\mathrm{E}^{+}$,
structure alignment encoder $\mathrm{E}^{*}$, and generator $\text{G}$
\State Initialize $\mathrm{E}^{+}$, $\mathrm{E}^{*}$, and $\text{G}$
\State Extract seen-class visual features $x_i=F(I_i)$ and compute class prototypes $p_c=\frac{1}{N_c}\sum_{i:y_i=c}x_i$
\State Estimate the dominant common direction $p_{\mathrm{com}}$
and refine the prototypes as
$\hat p_c=p_c-\alpha p_{\mathrm{com}}$
\State Compute and normalize attribute discriminability
$\{\bar d_j\}_{j=1}^{|A|}$ from the seen-class attribute matrix
\For{$t=1$ to $T$}
    \State Sample a mini-batch
    $\mathcal{B}\subset\mathcal{D}_{\mathrm{tr}}^{s}$
    \ForAll{$(x^{s},a^{s},y^{s})\in\mathcal{B}$}
        \State Predict attribute-specific parameters
        $\{(\mu_c^j,\sigma_c^j)\}_{j=1}^{|A|}\gets\mathrm{E}^{+}(a_c)$
        \State Compute the calibrated scale
        $\tilde{\sigma}_c^j=(1-\bar d_j)\sigma_c^j$
        \State Sample attribute residuals
        $\Delta a_c^j\sim\mathcal{N}(\mu_c^j,(\tilde{\sigma}_c^j)^2)$
        \State Construct the adaptive attribute
        $\tilde a_c^j\gets a_c^j+\lambda\Delta a_c^j$
        \State Obtain the visual-grounded embedding
        $v_c\gets\mathrm{E}^{*}(\tilde a_c)$
        \State Generate visual features using the conditional VAE-GAN:
        $\hat{x}\gets G(x^s,[\tilde a_c,v_c])$
    \EndFor
    \State Compute the generator loss $\mathcal{L}_{G}$, semantic consistency loss
    $\mathcal{L}_{SC}$, and prototype alignment loss $\mathcal{L}_{PA}$
    \State Compute
    $\mathcal{L}_{\mathrm{total}}\gets\mathcal{L}_{G}+\lambda_{SC}\mathcal{L}_{SC}+\lambda_{PA}\mathcal{L}_{PA}$
    \State Update $\mathrm{E}^{+}$, $\mathrm{E}^{*}$, and $\text{G}$ by minimizing
    $\mathcal{L}_{\mathrm{total}}$
\EndFor
\State \Return $\mathrm{E}^{+}$, $\mathrm{E}^{*}$, and $G$
\end{algorithmic}
\end{algorithm*}

\begin{algorithm*}[t]
\caption{Unseen-Class Feature Synthesis and ZSL Evaluation}
\label{alg:evaluate_zsl}
\begin{algorithmic}[1]
\Require Trained $\mathrm{E}^{+}$, $\mathrm{E}^{*}$, and $\text{G}$; unseen-class attributes
$\{a_u\}_{u\in\mathcal{Y}^{u}}$; seen-class training features
$\mathcal{X}_{\mathrm{tr}}^{s}$; numbers of synthesized features $N_{syn}$ and evaluation mode
$\in\{\mathrm{CZSL},\mathrm{GZSL}\}$
\Ensure Classification performance on unseen classes or on seen and unseen classes
\State Initialize the synthetic unseen-class set
$\widetilde{\mathcal{D}}^{u}\gets\emptyset$
\For{each unseen class $u\in\mathcal{Y}^{u}$}
    \State Predict $\{(\mu_u^j,\sigma_u^j)\}_{j=1}^{|A|}\gets\mathrm{E}^{+}(a_u)$
    \State Calibrate the variation scale using
    $\tilde{\sigma}_u^j=(1-\bar d_j)\sigma_u^j$
    \For{$n=1$ to $N_{syn}$}
        \State Sample $\Delta a_u^j\sim\mathcal{N}(\mu_u^j,(\tilde{\sigma}_u^j)^2)$
        \State Construct $\tilde a_u^j\gets a_u^j+\lambda\Delta a_u^j$
        \State Compute $v_u\gets\mathrm{E}^{*}(\tilde a_u)$ and sample latent noise $z$
        \State Generate $\hat x_{u,n}\gets G\!\left(z;[\tilde a_u,v_u]\right)$
        \State Add $(\hat x_{u,n},u)$ to $\widetilde{\mathcal{D}}^{u}$
    \EndFor
\EndFor
\If{$\mathrm{mode}=\mathrm{CZSL}$}
    \State Set classifier training data
    $\mathcal{D}_{\mathrm{cls}}\gets\widetilde{\mathcal{D}}^{u}$
\Else
    \State Set classifier training data
    $\mathcal{D}_{\mathrm{cls}}\gets\mathcal{X}_{\mathrm{tr}}^{s}\cup\widetilde{\mathcal{D}}^{u}$
\EndIf
\State Train a classifier $f$ on $\mathcal{D}_{\mathrm{cls}}$
\If{$\mathrm{mode}=\mathrm{CZSL}$}
    \State Evaluate $f$ on unseen-class test data and report Acc
\Else
    \State Evaluate $f$ on seen and unseen test data and report U, S, and
    $\text{H}=\frac{2\times\text{S}\times\text{U}}{\text{S}+\text{U}}$
\EndIf
\end{algorithmic}
\end{algorithm*}

\end{document}